\pdfoutput=1
\documentclass[conference]{IEEEtran}
\usepackage{cite}
\usepackage{amsmath,amssymb,amsfonts}
\usepackage{amsthm}
\usepackage{algorithmic}
\usepackage{algorithm}
\usepackage{graphicx}
\usepackage{textcomp}
\usepackage{booktabs}
\usepackage{multirow}
\usepackage{url}

\newtheorem{proposition}{Proposition}
\begin{document}

\title{Margin-Gated Cascade Reranking:\\ Does Per-Query Depth Selection\\ Beat a Fixed Top-$k$?}

\author{\IEEEauthorblockN{Phuong Huu Vu Tran}
\IEEEauthorblockA{Vietnamese-German University\\
10425032@student.vgu.edu.vn}}

\maketitle

\begin{abstract}
Strong cross-encoder rerankers are expensive, which motivates applying them only to
as many candidates as each query needs. We study a simple training-free instance of
this idea, \emph{margin-gated cascade reranking} (MGCR): a cheap cross-encoder scores
50 candidates, and a strong cross-encoder re-sorts a top prefix whose length is set
per query by the first large gap in the cheap scores. We evaluate it on SciFact,
NFCorpus, and FiQA with lexical and dense first stages. At its default setting MGCR
uses $14$--$17\times$ fewer strong-reranker calls than reranking all 50 candidates,
with $4.5$--$5.0\times$ lower modelled reranking latency. The saving does not come from
per-query adaptation. In all six settings the default is equivalent to always
reranking the top three (paired equivalence within $\pm0.01$ nDCG@10); reassigning
MGCR's prefix lengths to random queries changes nDCG@10 by less than $0.007$, with none
of 24 tests significant after multiple-comparison correction; and at matched call
budgets MGCR is never detectably better than fixed top-$k$. On FiQA, where full strong
reranking helps clearly, the default MGCR loses $0.021$--$0.035$ nDCG@10 against it.
Finally, skipping the cheap reranker and applying the strong one directly to the
first-stage top-10, at similar latency, is worse than MGCR on two of three BM25
settings but better on FiQA with dense retrieval. We conclude that cheap-score margins
are too weak a signal for choosing reranking depth in this setup, and that shuffled-depth
and direct-reranking controls are needed before crediting savings to an adaptive gate.
\end{abstract}

\begin{IEEEkeywords}
information retrieval, reranking, cross-encoder, cascade ranking, efficiency,
adaptive computation
\end{IEEEkeywords}

\section{Introduction}
A common retrieval pipeline recalls candidates with an efficient first stage and
reorders them with a cross-encoder that jointly encodes the query and each
passage~\cite{nogueira2019passagererank,nogueira2019multistage}. Every query--passage
pair costs one transformer forward pass, so larger and more accurate rerankers are
usually applied to fewer candidates. How many candidates to admit to each stage is a
long-standing quality--latency trade-off~\cite{wang2011cascade,nogueira2019multistage},
and several works choose that number per
query~\cite{culpepper2016cutoff,meng2024rlt}.

\begin{figure*}[!t]
\centering
\includegraphics[width=\textwidth]{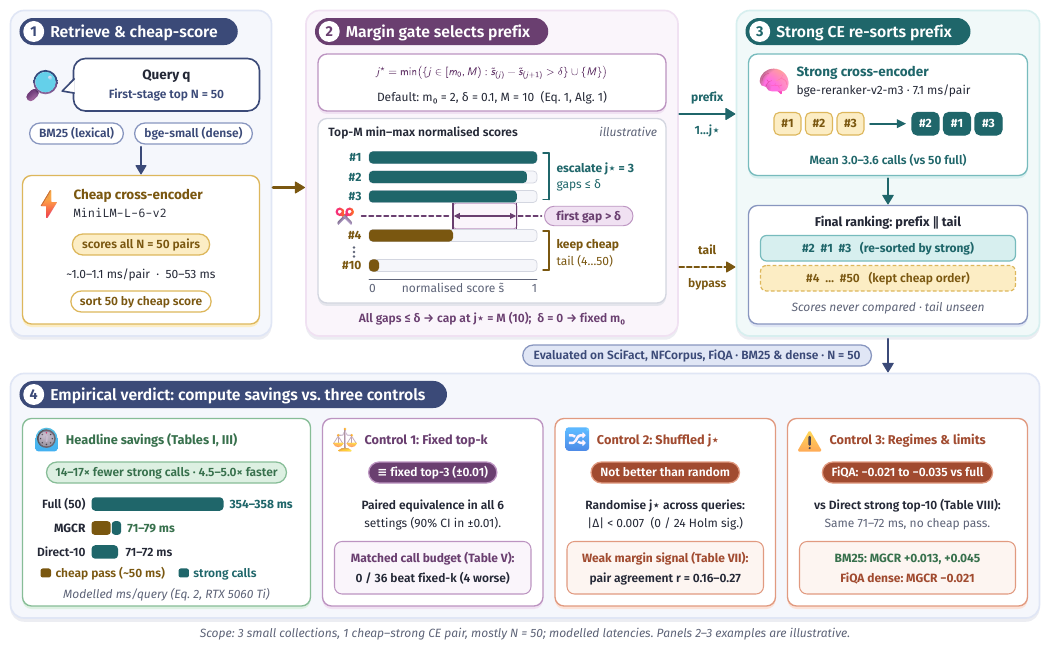}
\caption{Overview of MGCR and of the main findings. Panels 1--3 show the pipeline
(Eq.~\ref{eq:prefix}, Algorithm~\ref{alg:mgcr}); the example scores and ranks are
illustrative. Panel 4 summarises the evaluation, with modelled latencies
(Eq.~\ref{eq:cost}). Its short verdicts are statistical: in the matched-budget box,
``beat'' and ``worse'' mean a 95\% interval entirely above or below zero (uncorrected);
``not better than random'' means no significant difference after Holm correction, not
proof of no effect. The direct top-10 comparison is at similar latency; the two BM25
gains shown are for NFCorpus and FiQA, while that on SciFact ($+0.015$) is not
detectable.}
\label{fig:pipeline}
\end{figure*}

We examine an especially cheap way of doing so. If a cheap cross-encoder already
scores every candidate, the gaps between its top scores are available at no extra
cost. A large gap might mark a point below which the strong model need not look.
\emph{Margin-gated cascade reranking} (MGCR, Fig.~\ref{fig:pipeline}) implements this
rule. Showing that such a gate needs far fewer strong calls than full reranking is
easy, but a fixed top-$k$ cascade does that too. The question that matters is whether
the \emph{per-query} choice adds anything, and whether the cheap pass pays for itself.

\noindent\textbf{Contributions.}
(i) A precise description of MGCR and of what can and cannot be guaranteed about it
(Sec.~\ref{sec:method}).
(ii) An evaluation on three collections and two first stages with paired uncertainty
estimates (Sec.~\ref{sec:results}).
(iii) Three controls that isolate the value of the gate: matched-budget fixed top-$k$,
a shuffled-depth randomisation test, and strong reranking applied directly to the
first stage (Sec.~\ref{sec:controls}).
(iv) Diagnostics that show why the margin signal falls short, and an oracle that
indicates the room left for better depth predictors (Sec.~\ref{sec:analysis}).
The main finding is negative: the gate does not detectably improve on a fixed depth.

\section{Related Work}
\textbf{Cross-encoder rerankers.} Transformer
encoders~\cite{vaswani2017attention,devlin2019bert} enabled cross-encoder rerankers
based on BERT~\cite{nogueira2019passagererank,macavaney2019cedr} and on
sequence-to-sequence models~\cite{nogueira2020monot5,zhuang2023rankt5}. Large language
models have been used as listwise
rerankers~\cite{sun2023rankgpt,ma2023listwise,pradeep2023rankvicuna,pradeep2023rankzephyr}.
Late-interaction models reduce query-time cost by precomputing document token
embeddings~\cite{khattab2020colbert,santhanam2022colbertv2}.

\textbf{First-stage retrieval.} Candidates may come from lexical retrieval such as
BM25~\cite{robertson2009bm25}, from dense
retrievers~\cite{karpukhin2020dpr,reimers2019sbert,xiong2021ance,gao2021condenser,hofstatter2021tasb},
or from learned sparse models~\cite{formal2021splade,dai2019deepct}. Benchmarks
include MS MARCO~\cite{bajaj2016msmarco}, the TREC Deep Learning
tracks~\cite{craswell2020trecdl}, and BEIR~\cite{thakur2021beir}.

\textbf{Cascades, cutoffs, and adaptive reranking.} Cascade ranking learns or
configures sequences of increasingly expensive rankers that prune
candidates~\cite{wang2011cascade}, including multi-stage BERT
pipelines~\cite{nogueira2019multistage}. Candidates or computation can also be pruned
inside the transformer~\cite{soldaini2020cascade,xin2020early}, and precomputed
forward indexes with early stopping lower re-scoring cost~\cite{leonhardt2022forwardindexes}.
Ranked-list truncation chooses a per-query cutoff from score
distributions~\cite{arampatzis2009stop} or with learned models~\cite{bahri2020choppy}.
Closest to our question, per-query cutoffs have been predicted for multi-stage
retrieval~\cite{culpepper2016cutoff}, and truncation methods have been reproduced and
assessed as a way to set re-ranking depth~\cite{meng2024rlt}. Graph-based adaptive
re-ranking instead changes \emph{which} documents enter the pool within a
budget~\cite{macavaney2022gar}. MGCR is a heuristic, training-free member of the cutoff
family; we do not claim it as a new class of method, and we compare it only with fixed
depths and one score-level gate, not with learned truncation models.

\textbf{Difficulty and adaptive inference.} Query performance prediction estimates
how well retrieval will do for a query, for example from result-list language or score
statistics~\cite{cronen2002predicting,zhou2007qpp,shtok2012querydrift,carmel2010querydifficulty}.
Spending computation according to input difficulty also underlies adaptive computation
time and early-exit
inference~\cite{graves2016act,xin2020deebert,schwartz2020righttool,treviso2023efficient}.

\section{Method}
\label{sec:method}
For a query $q$, the first stage returns $N$ candidates. A cheap cross-encoder $f_c$
scores all of them, and they are sorted by that score. Let
$\tilde s_{(1)}\ge\dots\ge\tilde s_{(M)}$ be the top $M$ cheap scores after min--max
normalisation over those $M$ values (all zeros if they are equal). With a minimum
prefix $m_0$ and threshold $\delta\ge0$, MGCR selects
\begin{equation}
j^\star=\min\Big(\{\,j\in[m_0,M): \tilde s_{(j)}-\tilde s_{(j+1)}>\delta\}\cup\{M\}\Big),
\label{eq:prefix}
\end{equation}
re-sorts the first $j^\star$ candidates by the strong cross-encoder $f_s$, and places
them above the rest, which keep cheap order (Algorithm~\ref{alg:mgcr}). The default is
$m_0=2$, $\delta=0.1$, $M=10$. Setting $\delta=0$ gives fixed top-$m_0$ except when
scores tie exactly, and $\delta=1$ gives fixed top-$M$. Scores of the two models are
never compared with each other.

\begin{algorithm}[t]
\caption{MGCR for one query}
\label{alg:mgcr}
\begin{algorithmic}[1]
\STATE \textbf{Input:} candidates $P$, scorers $f_c,f_s$, threshold $\delta$, minimum
prefix $m_0$, cap $M$
\STATE sort $P$ by $f_c(q,\cdot)$, descending
\STATE $\tilde s\gets$ min--max normalisation of the top-$M$ cheap scores
\STATE $j\gets m_0$
\WHILE{$j<M$ \textbf{and} $\tilde s_{(j)}-\tilde s_{(j+1)}\le\delta$}
\STATE $j\gets j+1$
\ENDWHILE
\STATE re-sort $P_{1:j}$ by $f_s(q,\cdot)$
\STATE \textbf{return} $P_{1:j}\,\Vert\,P_{j+1:N}$ \quad (strong calls: $j$)
\end{algorithmic}
\end{algorithm}

\subsection{Two simple properties}
\begin{proposition}[Budget monotonicity]
For fixed scores, $j^\star$ is non-decreasing in $\delta$.
\end{proposition}
\begin{proof}
Every gap satisfying $\tilde s_{(j)}-\tilde s_{(j+1)}\le\delta$ also satisfies it for
$\delta'\ge\delta$, so the loop cannot stop earlier.
\end{proof}
\begin{proposition}[Top-$k$ exactness]
\label{prop:exact}
Let $A$ be the escalated prefix and $S_k$ the $k$ candidates with the highest strong
scores, assumed distinct. If $S_k\subseteq A$, MGCR's top $k$ equals, in order, that of
full strong reranking.
\end{proposition}
\begin{proof}
MGCR places $A$ first, ordered by $f_s$. The $k$ largest strong scores overall are all
in $A$, so they occupy positions $1,\dots,k$ in order.
\end{proof}
The condition is sufficient, not necessary, and it is restrictive: it requires
$k\le j^\star\le M$. Its empirical failure rate
$\pi_k=\Pr_q[S_k\not\subseteq A]$ is reported in Table~\ref{tab:diag}. At $k=10$ it is
$1.00$ in all six settings, so the proposition gives no useful guarantee for nDCG@10.

\subsection{Cost model}
Let $c_c$ and $c_s$ be the per-pair costs of the two models. Then
\begin{equation}
T_{\mathrm{MGCR}}=Nc_c+\mathbb{E}[j^\star]\,c_s,\qquad T_{\mathrm{full}}=Nc_s .
\label{eq:cost}
\end{equation}
Full strong reranking needs no cheap pass. A fixed top-$k$ cascade costs
$Nc_c+kc_s$, and strong reranking of the first-stage top $k$ costs $kc_s$. The ratio
$T_{\mathrm{full}}/T_{\mathrm{MGCR}}=\rho/(1+\rho\,\mathbb{E}[j^\star]/N)$, with
$\rho=c_s/c_c$, is therefore well below the call ratio $N/\mathbb{E}[j^\star]$.

\section{Experimental Setup}
\label{sec:setup}
\textbf{Data.} SciFact~\cite{wadden2020scifact}: 5{,}183 abstracts; we use the 188 of
300 development claims that have annotated evidence, with binary relevance (this is
not the BEIR test split). NFCorpus~\cite{boteva2016nfcorpus} in its BEIR
distribution~\cite{thakur2021beir}: 3{,}633 documents, 323 test queries, two relevance
grades. FiQA~\cite{maia2018fiqa} in its BEIR distribution: 57{,}638 documents, 648 test
queries, binary relevance. A document is its title and text (abstract for SciFact).

\textbf{Models.} First stage: Okapi BM25~\cite{robertson2009bm25} (the
\texttt{rank\_bm25} package with default parameters, lower-cased alphanumeric tokens,
no stemming or stop-word removal) or exact cosine search with
\texttt{bge-small-en-v1.5}~\cite{xiao2024cpack}, using its query instruction. Cheap
reranker: \texttt{ms-marco-MiniLM-L-6-v2}, a MiniLM~\cite{wang2020minilm} cross-encoder
from the Sentence-Transformers project~\cite{reimers2019sbert} trained on MS
MARCO~\cite{bajaj2016msmarco}. Strong reranker:
\texttt{bge-reranker-v2-m3}~\cite{chen2024m3}. Both rerankers run in half precision with
inputs truncated to 384 tokens. $N=50$ unless stated.

\textbf{Metrics and statistics.} nDCG@10~\cite{jarvelin2002ndcg} with linear gain,
normalised by all judged relevant documents, and Recall@10. Differences are paired by
query, with 95\% percentile bootstrap intervals (10{,}000 resamples) and two-sided
Wilcoxon signed-rank tests~\cite{wilcoxon1945} on non-zero differences. Equivalence is
declared when the 90\% bootstrap interval lies inside $\pm\varepsilon$, the interval
form of two one-sided tests~\cite{schuirmann1987}. The Holm
procedure~\cite{holm1979} corrects for multiplicity within a stated family.

\textbf{Latency.} Latencies are \emph{modelled}, not measured end to end: they combine
Eq.~\ref{eq:cost} with per-pair times recorded while scoring batches of 50 pairs,
tokenisation included, on one RTX~5060~Ti (at $N=50$, about 1.0--1.1\,ms cheap and 7.1\,ms strong).
They exclude first-stage retrieval and ignore any change in per-pair cost for the
smaller batches of a gated pipeline.

\textbf{Protocol.} There is no held-out tuning split. The grid of $\delta$, $m_0$, and
$M$ was explored on evaluation data, including an initial 50-query subset of the SciFact
development claims, so all settings should be read as in-sample; we report the sweeps
(Table~\ref{tab:sweep}). The controls of Sec.~\ref{sec:controls} and the equivalence
margins were chosen after the first results had been seen.

\section{Main Results}
\label{sec:results}
\begin{table*}[!t]
\caption{nDCG@10 at $N=50$. ``Fixed-3'' reranks the cheap top three with the strong
model. MGCR uses $\delta=0.1$, $M=10$; its mean strong calls per query are in
parentheses (full reranking uses 50).}
\label{tab:main}
\centering
\begin{tabular}{llrccccc}
\toprule
Data & 1st stage & $|Q|$ & 1st only & cheap & fixed-3 & MGCR (calls) & full \\
\midrule
SciFact & BM25 & 188 & 0.812 & 0.850 & 0.867 & 0.871 (3.5) & 0.867 \\
SciFact & dense & 188 & 0.860 & 0.874 & 0.885 & 0.883 (3.6) & 0.889 \\
NFCorpus & BM25 & 323 & 0.306 & 0.338 & 0.338 & 0.336 (3.2) & 0.336 \\
NFCorpus & dense & 323 & 0.344 & 0.351 & 0.349 & 0.349 (3.3) & 0.350 \\
FiQA & BM25 & 648 & 0.218 & 0.304 & 0.317 & 0.317 (3.1) & 0.338 \\
FiQA & dense & 648 & 0.403 & 0.383 & 0.398 & 0.398 (3.0) & 0.434 \\
\bottomrule
\end{tabular}

\end{table*}

\begin{table*}[!t]
\caption{Paired nDCG@10 differences with 95\% bootstrap intervals. $^{\ast}$: Wilcoxon
$p<0.05$ after Holm correction over these 18 tests. Help/hurt: share of queries where
full reranking raises/lowers nDCG@10 relative to the cheap ranking.}
\label{tab:paired}
\centering
\footnotesize
\begin{tabular}{llcccc}
\toprule
Data & 1st stage & full $-$ cheap & MGCR $-$ full & MGCR $-$ fixed-3 & help/hurt (\%) \\
\midrule
SciFact & BM25 & +0.017 [$-$0.008, +0.043] & +0.004 [$-$0.017, +0.025] & +0.004 [$-$0.001, +0.010] & 13/9 \\
SciFact & dense & +0.015 [$-$0.012, +0.043] & $-$0.006 [$-$0.027, +0.014] & $-$0.002 [$-$0.006, +0.003] & 14/10 \\
NFCorpus & BM25 & $-$0.002 [$-$0.014, +0.009] & 0.000 [$-$0.010, +0.009] & $-$0.002 [$-$0.005, +0.001] & 25/26 \\
NFCorpus & dense & $-$0.001 [$-$0.013, +0.011] & $-$0.001 [$-$0.011, +0.008] & $-$0.001 [$-$0.003, +0.002] & 30/26 \\
FiQA & BM25 & +0.034 [+0.023, +0.045]$^{\ast}$ & $-$0.021 [$-$0.030, $-$0.012]$^{\ast}$ & 0.000 [$-$0.005, +0.004] & 24/11 \\
FiQA & dense & +0.051 [+0.036, +0.065]$^{\ast}$ & $-$0.035 [$-$0.048, $-$0.023]$^{\ast}$ & 0.000 [$-$0.005, +0.005] & 37/18 \\
\bottomrule
\end{tabular}

\end{table*}

Tables~\ref{tab:main} and~\ref{tab:paired} give the main comparison.

\textbf{Where does the strong model help?} Only on FiQA is full strong reranking
detectably better than the cheap reranker ($+0.034$ with BM25, $+0.051$ with dense
retrieval). On SciFact the point estimates are positive ($+0.017$, $+0.015$) but the
intervals include zero; with 188 queries a gain of this size cannot be resolved. On
NFCorpus the difference is near zero. This does not mean reranking is useless there:
with BM25, both rerankers improve on the first stage (full $0.336$ and cheap $0.338$
against $0.306$; both intervals exclude zero). Averages also hide offsetting effects:
on NFCorpus full reranking changes many queries in both directions (help/hurt
25/26\% and 30/26\%).

\textbf{Cheap reranking can hurt.} On FiQA with dense retrieval the MiniLM reranker
scores \emph{below} the first stage ($0.383$ against $0.403$; interval of the
difference excludes zero). The default MGCR ($0.398$) does not detectably improve on
that first stage.

\textbf{Cost.} The default MGCR averages 3.0--3.6 strong calls, $14$--$17\times$ fewer
than full reranking. Because the cheap model must score all 50 candidates, the
modelled latency ratio is smaller, $4.5$--$5.0\times$
(Table~\ref{tab:latency}). A fixed-3 cascade is about as fast or slightly faster.

\begin{table}[!htb]
\caption{Modelled reranking latency in ms per query at $N=50$ (Eq.~\ref{eq:cost}).}
\label{tab:latency}
\centering
\footnotesize
\setlength{\tabcolsep}{4pt}
\begin{tabular}{llccccc}
\toprule
Data & 1st stage & cheap & fixed-3 & MGCR & full & full\,/\,MGCR \\
\midrule
SciFact & BM25 & 53 & 75 & 79 & 357 & 4.5$\times$ \\
SciFact & dense & 52 & 74 & 78 & 358 & 4.6$\times$ \\
NFCorpus & BM25 & 52 & 73 & 74 & 358 & 4.8$\times$ \\
NFCorpus & dense & 50 & 72 & 74 & 356 & 4.8$\times$ \\
FiQA & BM25 & 50 & 72 & 72 & 354 & 4.9$\times$ \\
FiQA & dense & 50 & 71 & 71 & 355 & 5.0$\times$ \\
\bottomrule
\end{tabular}

\end{table}

\textbf{Quality against full reranking.} On FiQA the default MGCR is clearly worse
($-0.021$, $-0.035$). On SciFact and NFCorpus no difference is detected. On SciFact this
is not evidence of equivalence: the 90\% intervals ($[-0.014,+0.021]$ and
$[-0.023,+0.011]$) are too wide for a $\pm0.02$ margin on both first stages, a margin
that would in any case exceed the observed gain of full over cheap reranking. On
NFCorpus the 90\% intervals lie within $\pm0.01$.

\textbf{Quality against fixed-3.} Here equivalence does hold. All six 90\% intervals of
MGCR minus fixed-3 lie within $\pm0.01$ (all endpoints are between $-0.006$
and $+0.010$). In terms of nDCG@10, the default gate behaves
like a constant depth of three.

\section{Does the Gate Add Value?}
\label{sec:controls}
\begin{figure*}[t]
\centering
\includegraphics[width=\textwidth]{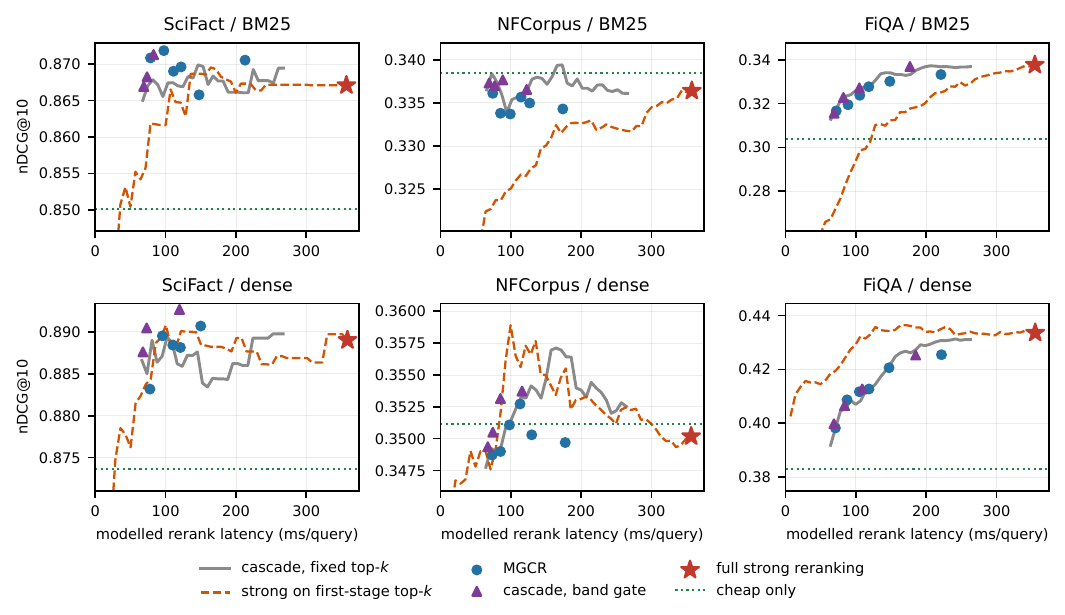}
\caption{nDCG@10 against modelled latency. Cascade methods (fixed top $k\le30$ for the line) pay about 50\,ms for the
cheap pass before the first strong call; direct strong reranking of the first-stage
top $k$ ($k\le50$) does not. Vertical ranges are chosen per panel, and some very
shallow direct points fall below them. The vertical scales differ by an order of
magnitude between collections.}
\label{fig:frontier}
\end{figure*}

\begin{table}[!htb]
\caption{Raising the cap on FiQA ($N=50$). ``Minus full'': paired nDCG@10 difference
from full strong reranking, with a 95\% interval.}
\label{tab:cap}
\centering
\resizebox{\columnwidth}{!}{
\begin{tabular}{llcccc}
\toprule
1st stage & Method ($\delta{=}0.2$) & nDCG@10 & calls & ms & minus full \\
\midrule
BM25 & MGCR, $M{=}10$ & 0.319 & 5.4 & 89 & $-$0.018 [$-$0.026, $-$0.011] \\
BM25 & MGCR, $M{=}20$ & 0.330 & 13.8 & 148 & $-$0.008 [$-$0.014, $-$0.002] \\
BM25 & MGCR, $M{=}30$ & 0.333 & 24.1 & 221 & $-$0.004 [$-$0.009, 0.000] \\
BM25 & MGCR, $M{=}50$ & 0.337 & 45.9 & 376 & $-$0.001 [$-$0.002, 0.000] \\
BM25 & fixed-20 & 0.336 & 20.0 & 192 & $-$0.002 [$-$0.005, +0.001] \\
BM25 & fixed-30 & 0.337 & 30.0 & 263 & $-$0.001 [$-$0.004, +0.001] \\
BM25 & full & 0.338 & 50.0 & 354 & --- \\
\midrule
dense & MGCR, $M{=}10$ & 0.409 & 5.3 & 88 & $-$0.025 [$-$0.036, $-$0.014] \\
dense & MGCR, $M{=}20$ & 0.421 & 13.7 & 147 & $-$0.013 [$-$0.021, $-$0.006] \\
dense & MGCR, $M{=}30$ & 0.425 & 24.2 & 222 & $-$0.008 [$-$0.014, $-$0.003] \\
dense & MGCR, $M{=}50$ & 0.433 & 46.4 & 379 & $-$0.001 [$-$0.003, 0.000] \\
dense & fixed-20 & 0.429 & 20.0 & 192 & $-$0.004 [$-$0.009, 0.000] \\
dense & fixed-30 & 0.431 & 30.0 & 263 & $-$0.003 [$-$0.006, +0.001] \\
dense & full & 0.434 & 50.0 & 355 & --- \\
\bottomrule
\end{tabular}
}
\end{table}

\subsection{Matched strong-call budget}
Each MGCR operating point, $(\delta,M)\in\{0.1,0.2,0.3,0.5\}\times\{10\}\cup
\{(0.2,20),(0.2,30)\}$, has a mean budget $b$. We compare it, query by query, with the
mixture of fixed top-$\lfloor b\rfloor$ and top-$(\lfloor b\rfloor{+}1)$ that spends the
same $b$ on average. As a second adaptive rule we use a \emph{band gate} that escalates
every candidate whose cheap score, min--max normalised over all 50, is within
$\beta\in\{0.1,0.2,0.3,0.5\}$ of the top (at least two candidates).

\begin{table*}[!t]
\caption{Adaptive gates minus fixed top-$k$ at the same mean budget. ``$+/-$'' counts
operating points whose 95\% interval lies entirely above/below zero (not corrected for
multiplicity).}
\label{tab:matched}
\centering
\footnotesize
\begin{tabular}{llcccccc}
\toprule
 & & \multicolumn{3}{c}{MGCR (6 points)} & \multicolumn{3}{c}{band gate (4 points)} \\
\cmidrule(lr){3-5}\cmidrule(lr){6-8}
Data & 1st stage & mean $|\Delta|$ & range of $\Delta$ & $+$/$-$ & mean $|\Delta|$ & range of $\Delta$ & $+$/$-$ \\
\midrule
SciFact & BM25 & 0.0037 & [$-$0.004, +0.006] & 0/0 & 0.0038 & [+0.002, +0.009] & 1/0 \\
SciFact & dense & 0.0036 & [$-$0.004, +0.007] & 0/0 & 0.0057 & [+0.001, +0.009] & 0/0 \\
NFCorpus & BM25 & 0.0022 & [$-$0.005, 0.000] & 0/0 & 0.0008 & [$-$0.001, +0.002] & 0/0 \\
NFCorpus & dense & 0.0020 & [$-$0.007, 0.000] & 0/1 & 0.0017 & [+0.001, +0.004] & 1/0 \\
FiQA & BM25 & 0.0026 & [$-$0.004, $-$0.001] & 0/2 & 0.0012 & [0.000, +0.004] & 1/0 \\
FiQA & dense & 0.0021 & [$-$0.005, +0.004] & 0/1 & 0.0025 & [$-$0.002, +0.004] & 0/0 \\
\bottomrule
\end{tabular}

\end{table*}

Table~\ref{tab:matched} summarises the result. Of 36 MGCR points, none is detectably
above fixed top-$k$ and four are below it, with all gaps within $\pm0.0075$. The band
gate is detectably above in 3 of 24 points and never below, with gaps of at most
$0.009$; given the uncorrected counts and small sizes we regard this as inconclusive.
Neither rule shows a consistent advantage over a constant depth
(Fig.~\ref{fig:frontier}).

\subsection{Shuffled-depth randomisation}
Interpolation compares MGCR with a different distribution of depths. A sharper control
keeps MGCR's exact multiset of $j^\star$ values and assigns them to queries at random.
Cost is unchanged; only the link between a query and its depth is destroyed. If margins
identify the queries that need depth, true MGCR should win. Table~\ref{tab:shuffle}
reports MGCR minus the expected nDCG@10 under random assignment, with a permutation
test (10{,}000 permutations). The 24 differences lie between $-0.003$ and $+0.006$, and
13 are positive. Two have uncorrected $p<0.05$, one in each direction, and neither
survives Holm correction (smallest adjusted $p=0.15$). We therefore find no reliable
evidence that the margin gate assigns depth better than chance. This does not prove the
effect is zero; the upper ends of the corresponding bootstrap intervals reach about
$0.013$.

\begin{table}[!htb]
\caption{MGCR minus the shuffled-depth control in nDCG@10, with the permutation
$p$-value in parentheses. $^{\dagger}$: uncorrected $p<0.05$.}
\label{tab:shuffle}
\centering
\resizebox{\columnwidth}{!}{
\begin{tabular}{llcccc}
\toprule
Data & 1st stage & $\delta{=}.1,M{=}10$ & $\delta{=}.2,M{=}10$ & $\delta{=}.2,M{=}20$ & $\delta{=}.2,M{=}30$ \\
\midrule
SciFact & BM25 & +0.005 (0.15) & +0.005 (0.15) & 0.000 (0.94) & +0.002 (0.65) \\
SciFact & dense & $-$0.003 (0.35) & +0.003 (0.47) & +0.004 (0.34) & +0.005 (0.28) \\
NFCorpus & BM25 & $-$0.001 (0.54) & $-$0.003$^{\dagger}$ (0.05) & $-$0.002 (0.38) & $-$0.002 (0.39) \\
NFCorpus & dense & 0.000 (0.98) & $-$0.001 (0.51) & $-$0.001 (0.66) & $-$0.001 (0.63) \\
FiQA & BM25 & +0.001 (0.71) & $-$0.001 (0.48) & +0.001 (0.68) & +0.001 (0.67) \\
FiQA & dense & +0.002 (0.24) & +0.006$^{\dagger}$ (0.006) & +0.003 (0.26) & +0.002 (0.51) \\
\bottomrule
\end{tabular}
}
\end{table}

\begin{table}[!htb]
\caption{Diagnostics. Agreement in the lowest and highest margin deciles; pair-level
correlation $r$; failure rate $\pi_k$ of the condition in
Proposition~\ref{prop:exact} at default settings; and the label-using oracle's nDCG@10
(mean strong calls).}
\label{tab:diag}
\centering
\resizebox{\columnwidth}{!}{
\begin{tabular}{llcccccc}
\toprule
 & & \multicolumn{3}{c}{margin vs.\ agreement} & \multicolumn{2}{c}{$\pi_k$} & \\
\cmidrule(lr){3-5}\cmidrule(lr){6-7}
Data & 1st stage & low & high & $r$ & $k{=}1$ & $k{=}3$ & oracle (calls) \\
\midrule
SciFact & BM25 & 0.48 & 0.92 & 0.26 & 0.10 & 0.85 & 0.891 (0.5) \\
SciFact & dense & 0.50 & 0.95 & 0.27 & 0.09 & 0.86 & 0.913 (0.5) \\
NFCorpus & BM25 & 0.35 & 0.80 & 0.21 & 0.35 & 0.89 & 0.353 (1.7) \\
NFCorpus & dense & 0.47 & 0.80 & 0.18 & 0.35 & 0.89 & 0.368 (2.1) \\
FiQA & BM25 & 0.49 & 0.80 & 0.18 & 0.30 & 0.89 & 0.345 (1.1) \\
FiQA & dense & 0.52 & 0.76 & 0.16 & 0.31 & 0.92 & 0.441 (1.9) \\
\bottomrule
\end{tabular}
}
\end{table}

\begin{figure}[!htb]
\centering
\includegraphics[width=\columnwidth]{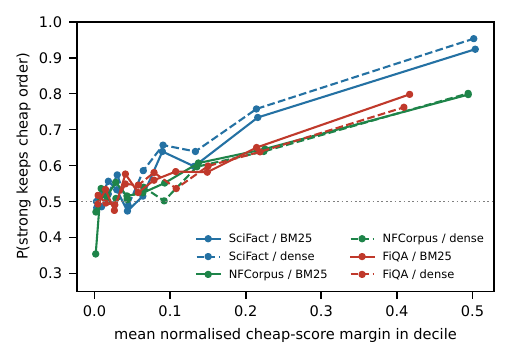}
\caption{Probability that the strong reranker keeps the cheap order of an adjacent
pair, by decile of the normalised cheap-score margin (cheap top-10).}
\label{fig:calib}
\end{figure}

\subsection{Is the cheap pass worth its latency?}
The cheap pass costs about as much as seven strong calls. The obvious alternative
spends that time on the strong model, applied directly to the first-stage top $k$.
Table~\ref{tab:direct} shows that $k=10$ takes 71--72\,ms against 71--79\,ms for MGCR.
At this budget MGCR is detectably better on NFCorpus/BM25 ($+0.013$) and FiQA/BM25
($+0.045$), detectably worse on FiQA/dense ($-0.021$), and not separable in the other
three settings. On FiQA/dense, direct reranking of the top 20 (142\,ms) reaches the
quality of full reranking ($0.434$), and even $k=3$ (21\,ms) has a higher mean than MGCR.
With 143\,ms, direct $k=20$ is within $0.002$ of full reranking on SciFact as well. The
value of the cascade thus depends on the first stage: it helped most where an untuned
BM25 left much for the cheap model to repair.

\begin{table*}[!t]
\caption{MGCR versus strong reranking of the first-stage top $k$ without a cheap pass:
nDCG@10 (modelled ms per query). The last column is paired, with a 95\% interval.}
\label{tab:direct}
\centering
\footnotesize
\begin{tabular}{llccccc}
\toprule
 & & & \multicolumn{3}{c}{strong on first-stage top-$k$ (no cheap pass)} & \\
\cmidrule(lr){4-6}
Data & 1st stage & MGCR & $k{=}3$ & $k{=}10$ & $k{=}20$ & MGCR $-$ ($k{=}10$) \\
\midrule
SciFact & BM25 & 0.871 (79) & 0.840 (21) & 0.856 (71) & 0.869 (143) & +0.015 [$-$0.010, +0.042] \\
SciFact & dense & 0.883 (78) & 0.866 (22) & 0.884 (72) & 0.890 (143) & $-$0.001 [$-$0.024, +0.023] \\
NFCorpus & BM25 & 0.336 (74) & 0.311 (21) & 0.323 (72) & 0.330 (143) & +0.013 [+0.003, +0.024] \\
NFCorpus & dense & 0.349 (74) & 0.347 (21) & 0.348 (71) & 0.355 (143) & +0.001 [$-$0.011, +0.014] \\
FiQA & BM25 & 0.317 (72) & 0.238 (21) & 0.272 (71) & 0.310 (142) & +0.045 [+0.029, +0.061] \\
FiQA & dense & 0.398 (71) & 0.413 (21) & 0.419 (71) & 0.434 (142) & $-$0.021 [$-$0.035, $-$0.008] \\
\bottomrule
\end{tabular}

\end{table*}

\begin{table*}[!t]
\caption{Sensitivity at $M=10$: nDCG@10 (mean strong calls).}
\label{tab:sweep}
\centering
\footnotesize
\setlength{\tabcolsep}{4.5pt}
\begin{tabular}{llccccccc}
\toprule
 & & \multicolumn{5}{c}{$\delta$ ($m_0{=}2$)} & \multicolumn{2}{c}{$m_0$ ($\delta{=}0.1$)} \\
\cmidrule(lr){3-7}\cmidrule(lr){8-9}
Data & 1st stage & 0 & 0.1 & 0.2 & 0.3 & 0.5 & 1 & 3 \\
\midrule
SciFact & BM25 & 0.865 (2.0) & 0.871 (3.5) & 0.872 (6.2) & 0.869 (8.1) & 0.870 (9.6) & 0.861 (0.3) & 0.870 (5.3) \\
SciFact & dense & 0.887 (2.0) & 0.883 (3.6) & 0.890 (6.1) & 0.888 (8.1) & 0.888 (9.6) & 0.885 (0.3) & 0.887 (5.4) \\
NFCorpus & BM25 & 0.337 (2.0) & 0.336 (3.2) & 0.334 (4.7) & 0.334 (6.7) & 0.336 (8.8) & 0.338 (0.8) & 0.336 (5.2) \\
NFCorpus & dense & 0.348 (2.0) & 0.349 (3.3) & 0.349 (4.9) & 0.351 (6.7) & 0.353 (8.8) & 0.351 (0.7) & 0.351 (5.2) \\
FiQA & BM25 & 0.313 (2.0) & 0.317 (3.1) & 0.319 (5.4) & 0.324 (7.8) & 0.328 (9.6) & 0.309 (0.5) & 0.321 (4.7) \\
FiQA & dense & 0.392 (2.0) & 0.398 (3.0) & 0.409 (5.3) & 0.412 (7.8) & 0.413 (9.7) & 0.391 (0.5) & 0.403 (4.6) \\
\bottomrule
\end{tabular}

\end{table*}

\begin{table}[!htb]
\caption{Candidate depth $N$ (SciFact, BM25).}
\label{tab:depth}
\centering
\footnotesize
\setlength{\tabcolsep}{3.5pt}
\begin{tabular}{rccccccc}
\toprule
 & \multicolumn{3}{c}{nDCG@10} & MGCR & \multicolumn{2}{c}{modelled ms} & \\
\cmidrule(lr){2-4}\cmidrule(lr){6-7}
$N$ & full & fixed-3 & MGCR & calls & full & MGCR & ratio \\
\midrule
20 & 0.869 & 0.874 & 0.868 & 3.49 & 143 & 51 & 2.8$\times$ \\
50 & 0.867 & 0.867 & 0.871 & 3.54 & 357 & 79 & 4.5$\times$ \\
100 & 0.868 & 0.868 & 0.870 & 3.48 & 717 & 128 & 5.6$\times$ \\
\bottomrule
\end{tabular}

\end{table}

\subsection{Deeper reranking on FiQA}
Because FiQA rewards depth, one can raise both $\delta$ and $M$
(Table~\ref{tab:cap}). At $\delta=0.2$, $M=20$ recovers 78\% (BM25) and 74\% (dense) of
the gain of full over cheap reranking, and $M=30$ recovers 87\% and 84\%, still below
full reranking. A constant depth of 20 has a higher mean nDCG@10 than MGCR with
$M=30$ ($0.336$ against $0.333$, and $0.429$ against $0.425$) with fewer strong calls
and less time.

\section{Analysis}
\label{sec:analysis}
\textbf{Margins are informative only at the top end.} For adjacent pairs in each cheap
top-10, Fig.~\ref{fig:calib} shows how often the strong model keeps the cheap order, by
decile of normalised margin. In the lower six deciles agreement stays between $0.35$
and $0.59$ with no consistent trend; it rises mainly in the top three, reaching
$0.76$--$0.95$ in the highest decile. The pair-level correlation between margin and
agreement is only $r=0.16$--$0.27$ (Table~\ref{tab:diag}). This analysis is
descriptive: it does not control for rank position, and agreement on adjacent pairs is
not the same as knowing how deep relevant documents lie.

\textbf{Disagreement is not gain.} One might guess how much a collection benefits from
the strong model by how often the two models disagree, which needs no labels. In our
data this fails: they choose a different top document for 55\% of NFCorpus queries,
where no average gain is seen, and for 53--55\% of FiQA queries, where the gain is
largest.

\textbf{Oracle depth.} An oracle chooses per query the prefix length in
$\{0\}\cup\{2,\dots,10\}$ that maximises nDCG@10, taking the shortest on ties. It
reaches $0.891$ on SciFact/BM25 and $0.441$ on FiQA/dense, the latter above full
reranking ($0.434$), with at most 2.1 mean calls in any setting. For 65--87\% of queries
no prefix improves on the cheap ranking. Because it selects among ten rankings using
the test labels, this is an optimistic ceiling and says nothing about how predictable
the best depth is. It does show that the negative result concerns this predictor, not
necessarily per-query depth in general.

\textbf{Sensitivity.} Table~\ref{tab:sweep} varies $\delta$ and $m_0$. Larger $\delta$
spends more calls; quality rises steadily on FiQA and barely moves elsewhere. No value
of $m_0$ is best everywhere: $m_0=2$ has the highest mean on SciFact/BM25, while
$m_0=3$ is higher on both FiQA settings at greater cost.

\textbf{Candidate depth.} Table~\ref{tab:depth} varies $N$ on SciFact/BM25. MGCR's
strong calls stay near 3.5, as expected since $M=10$ bounds them (a fixed depth is flat
by definition), so the latency ratio to full reranking grows from $2.8\times$ to
$5.6\times$. The cheap pass grows linearly with $N$ in the cascade.

\textbf{Two selected queries.} These illustrate opposite cases and are not
representative samples. For the SciFact claim beginning ``1,000 genomes project
enables mapping\ldots'' (BM25), the cheap model ranks the annotated abstract first; full
reranking moves it to sixth (nDCG@10 $1.00\to0.36$), whereas MGCR, like fixed-3, keeps
it first. For the FiQA query ``Non Resident aliens -- Question of standard vs
itemized'', the only judged passage is seventh in the cheap ranking. The default gate
stops before reaching it (nDCG@10 $0.33$), while fixed top-10 and full reranking move it
to first.

\section{Limitations}
\label{sec:limits}
The conclusions are specific to one cheap--strong model pair, three small collections,
and mostly $N=50$; they should not be extrapolated to web-scale benchmarks, LLM
rerankers, or other candidate depths. SciFact has few queries, so small effects there
are unresolved. Hyperparameters were not tuned on held-out data, several analyses were
added after initial results, and only some test families are corrected for
multiplicity. Absence of a significant difference is not evidence of equivalence except
where an equivalence interval is reported. The BM25 baseline is untuned, which may
flatter any reranker applied to it. Latencies are modelled from batch throughput on a
single GPU. We did not compare with learned truncation or query performance prediction
methods, which are the natural next test, ideally against the same shuffled-depth and
direct-reranking controls.

\section{Conclusion}
Margin-gated cascade reranking saves most strong-reranker calls relative to full
reranking, but in our experiments a constant shallow depth saves them equally well. The
score-gap gate was not detectably better than random assignment of its own depths, lost
quality where deep reranking mattered, and did not always justify the cheap pass it
relies on. Claims for adaptive reranking depth should be tested against a fixed depth at
equal budget, a shuffled-depth control, and direct strong reranking at equal latency.

\section*{Reproducibility}
The ancillary files distributed with this preprint contain the code, the candidate lists with both rerankers'
scores for every query, the run metadata with per-pair timings, and a script that
regenerates all tables and figures on a CPU. The datasets are public and are not
redistributed. As a check, we rebuilt all candidate lists and re-scored 8 random
queries per setting on a CPU in full precision. BM25 lists matched exactly; dense
lists contained the same documents, in identical order for at least $98.7\%$ of
queries. Scores differed from the saved half-precision values by at most $0.08$ logits,
and the strong model's top-10 order was identical for 45 of the 48 queries.

\bibliographystyle{IEEEtran}
\bibliography{references}

\end{document}